# A Linear Belief Function Approach to Portfolio Evaluation

**Liping Liu**
Department of Management
University of Akron
Akron, OH 44325-4801

**Catherine Shenoy**
University of Kansas
School of Business
Lawrence, KS 66045-7585

**Prakash P. Shenoy**
University of Kansas
School of Business
Lawrence, KS 66045-7585

## Abstract

We show how to use linear belief functions to represent market information and financial knowledge, including complete ignorance, statistical observations, subjective speculations, distributional assumptions, linear relations, and empirical asset pricing models. We then appeal to Dempster's rule of combination to integrate the knowledge for assessing an overall belief on portfolio performance, and to update this belief by incorporating additional information.

## 1 INTRODUCTION

In modern portfolio analysis, a financial asset is characterized as a random variable with a probability distribution over its possible returns. A portfolio is a linear combination of asset variables (Tobin 1959) and hence a random variable itself with a return distribution functionally determined by the return distributions of the individual assets. With such an abstraction, the problem of selecting an optimal portfolio becomes one of ranking probability distributions by their mean-variance positions (Markowitz 1959), general moments (Samuelson 1970), or coarse approximations (Liu 1999a).

Since return distributions carry sufficient information for evaluating expected returns and risks, their determination is crucial for the practice of portfolio management and financial decision-making. In the finance literature, there has been ample evidence suggesting that return distributions are predictable (Campbell 2000). However, one often finds it difficult to translate this evidence of predictability into practical portfolio advice (Aït-Sahalia and Brandt 2001). Among many questions to be addressed to fill in the gap, an important one is how to integrate a variety of information concerning asset returns to update the return distribution. At least two approaches to portfolio selection are commonly used in finance (Pástor 2000). A "data-based" approach assumes a functional form for a return distribution and estimates its parameters from the time series of returns. For example, sample estimates of mean and variance may be used to compute the optimal portfolio in a mean-variance framework. This approach ignores potential usefulness of various asset pricing models, including the Capital Asset Pricing Model (CAPM) of Sharpe (1964) and Lintner (1965), the arbitrage pricing theory of Ross (1976), as well as their multifactor extensions (Fama 1996). On the other hand, in an approach based on asset pricing models, the optimal portfolio is a combination of benchmark portfolios that expose an investor only to priced sources of risk. For example, the CAPM-based approach prescribes the market portfolio as a single optimal portfolio for every investor. This model-based approach does not use historical data on non-benchmark assets. These two typical approaches essentially reflect two extreme positions of using information for portfolio evaluation; the first approach bases on historical data while the second one bases on finance theory. The portfolio literature is silent about what happens in between except for a few recent attempts in combining sample evidence with the CAPM in a Bayesian framework (Baks et al. 2001, Pástor 2000).

Besides the selective use of historical data and theoretical models, the finance literature overlooks a third category of information, which may be called *soft data*, for portfolio evaluation. Soft data includes corporate and government announcements, news reports on industry and political events, as well as subjective judgments. The finance literature generally leaves the usage of soft data to financial analysts and individual speculators except that Shenoy and Shenoy (1999, 2002) model them using Bayesian probabilities and discrete belief functions.

The goal of this study is to propose a linear belief function (LBF) approach to portfolio evaluation. We use LBFs to represent the knowledge available for determining a portfolio distribution, including historical data, subjective beliefs, observations, theoretical linear relations, empirical linear regression models, and ignorance. We integrate the knowledge using Dempster's rule of combination. We illustrate the approach using an example of three gold-mining stocks. We also illustrate dynamic belief updating by entering additional information.

## 2 LINEAR BELIEF FUNCTIONS

The Dempster-Shafer theory of belief functions (Shafer 1976) provides a flexible tool for knowledge representation and a rigorous mechanism for knowledge integration. The notion of LBFs extends the theory to the case when variables of interest are continuous (Dempster 1990, Liu 1996). In particular, a LBF can represent both logical and probabilistic knowledge for three types of variables: deterministic such as observed random variables, random whose distribution is normal, and vacuous or complete ignorance. Logical knowledge is represented by linear equations, or geometrically, a hyperplane. Probabilistic knowledge is represented by a normal distribution.

Intuitively, a LBF represents knowledge regarding the location of the true value of variables as follows. We are certain that the truth is on a so-called *certainty hyperplane* but we do not know its exact location. In some dimensions of the certainty hyperplane, we believe the true value could be anywhere from $-\infty$ to $+\infty$ and the probability of being at a particular location is described by a normal distribution. In other dimensions, our knowledge is vacuous, i.e., the true value is somewhere from $-\infty$ to $+\infty$ but the associated probability is unknown. Technically, a LBF is a continuous analog of the basic probability assignment in the discrete case, in the sense that its focal elements are exclusive, parallel sub-hyperplanes over the certainty hyperplane and the belief mass is expressed as a normal distribution over the sub-hyperplanes (Liu 1996).

### 2.1 KNOWLEDGE REPRESENTATION

Shafer (1992) and Liu (1996) propose two mathematical representations of a LBF: a wide-sense inner product and a linear functional in the variable space, and as their duals over a hyperplane in the sample space. Monney (2003) proposes yet another structure called "Gaussian hints." Although these representations are mathematically elegant, they do not lend themselves to easy computation. As Liu (1996) shows, the two basic operations of LBFs cannot be done using one coherent representation; complex transformations must be performed in order to interplay the two operations. Dempster (1990) proposes representing a LBF indirectly by representing each of its degenerate components. In this sub-section, we re-elaborate on Dempster (1990) and present an elementary approach to knowledge representation.

The key to Dempster's approach is to use moment matrices and their transformations, called *sweepings* (Liu 1999b), to represent each non-trivial case of LBFs. In particular, assume $X$ is a row vector of normal variables with mean vector $\mu$ and covariance matrix $\Sigma$. Then, the multivariate normal distribution can be equivalently represented as a moment matrix:

$$M(X) = \begin{bmatrix} \mu \\ \Sigma \end{bmatrix}. \tag{1}$$

If the distribution is non-degenerate and $\Sigma$ has a full rank, the moment matrix can be turned into a fully swept form:

$$M(\vec{X}) = \begin{bmatrix} \mu\Sigma^{-1} \\ -\Sigma^{-1} \end{bmatrix}. \tag{2}$$

Note that, except for normalization, Equation (2) completely determines the normal density function for $X$. Thus, it represents the probability distribution of $X$ in the potential form (Lauritzen and Spiegelhalter 1988).

These simple matrices allow us to represent three special cases of LBFs. First, for an ordinary normal distribution, say $X \sim N(\mu, \Sigma)$, $M(X)$ in Equation (1) represents it. Second, suppose one makes a direct observation on $X$ and obtains a value $x$. In this case, since there is no uncertainty, both variance and covariance vanish, i.e., $\Sigma = 0$. Thus, a direct observation can be represented as:

$$M(X) = \begin{bmatrix} x \\ 0 \end{bmatrix}. \tag{3}$$

Third, suppose one is completely ignorant about $X$. In the Dempster-Shafer theory, one represents it by assigning the whole belief mass to the entire sample space. How do we represent the case as a LBF? The trick is to imagine complete ignorance as the limiting case when the variance of $X$ approaches to $\infty$. In this limiting case, it can be shown that $\Sigma^{-1} = 0$. Thus, by using the fully swept moment matrix in Equation (2), we have a finite representation of the vacuous LBF:

$$M(\vec{X}) = \begin{bmatrix} 0 \\ 0 \end{bmatrix}. \tag{4}$$

To represent the other special cases of a LBF, we need partial sweepings. Unlike a full sweeping, a partial sweeping is a transformation on a subset of variables. Suppose $X$ and $Y$ are two vectors of normal variables with mean vectors $\mu_i$ and covariance matrices $\Sigma_{ij}$, $i,j = 1, 2$. Then the joint moment matrix of $X$ and $Y$ is:

$$M(X,Y) = \begin{bmatrix} \mu_1 & \mu_2 \\ \Sigma_{11} & \Sigma_{12} \\ \Sigma_{21} & \Sigma_{22} \end{bmatrix}. \tag{5}$$

This matrix may be partially swept on any individual variable or a subset of variables in $X \cup Y$. Without loss of generality, we define its partial sweeping on $X$ as follows:

$$M(\vec{X},Y) = \begin{bmatrix} \mu_1(\Sigma_{11})^{-1} & \mu_2 - \mu_1(\Sigma_{11})^{-1}\Sigma_{12} \\ -(\Sigma_{11})^{-1} & (\Sigma_{11})^{-1}\Sigma_{12} \\ \Sigma_{21}(\Sigma_{11})^{-1} & \Sigma_{22} - \Sigma_{21}(\Sigma_{11})^{-1}\Sigma_{12} \end{bmatrix}. \tag{6}$$

From Equation (6) we can make two observations. First, after the partial sweeping on $X$, the mean vector and co-variance matrix of $X$ become $\mu_1(\Sigma_{11})^{-1}$ and $-(\Sigma_{11})^{-1}$, respectively. Compared to Equation (2), we see that the result is the same as that of a full sweeping of the marginal moment matrix of $X$. In other words, if we restrict the moment matrix $M(X,Y)$ in Equation (5) to $X$ indices and perform a full sweeping on $X$, then the mean vector and covariance matrix will be also transformed into $\mu_1(\Sigma_{11})^{-1}$ and $-(\Sigma_{11})^{-1}$, respectively. Thus, the elements corresponding to $X$ in Equation (6) represent the potential distribution of $X$. Second, according to multivariate statistics, $\mu_2 - \mu_1(\Sigma_{11})^{-1}\Sigma_{12}$ is the conditional mean of $Y$ given $X = 0$; $\Sigma_{22} - \Sigma_{21}(\Sigma_{11})^{-1}\Sigma_{12}$ is the conditional covariance of $Y$ given $X = 0$; and $(\Sigma_{11})^{-1}\Sigma_{12}$ is the regression coefficient of $Y$ on $X$ (Whittaker 1990 p. 162-163). Therefore, the elements corresponding to $Y$ indices and the intersection of $X$ and $Y$ in $M(\vec{X},Y)$ represents the conditional probability distribution of $Y$ given $X = 0$.

These semantics render the partial sweeping operation an efficient method for manipulating multivariate normal distributions. They also form the foundation of the moment matrix representations for the three remaining important cases of LBFs, including proper belief functions, linear equations, and linear regression models.

**Proper Linear Belief Functions**: Assume there exists a piece of evidence justifying a normal distribution for variables in $Y$ while bearing no-opinions for variables in $X$. Also, assume that $X$ and $Y$ are not perfectly linearly related, i.e., their correlation is less than 1. It is easy to see that this case involves a mix of an ordinary normal distribution for $Y$ and a vacuous belief function for $X$. Thus, we need to employ a partially swept matrix $M(\vec{X},Y)$. To represent the ignorance on $X$, we use the potential form of $X$ and set $\mu_1(\Sigma_{11})^{-1} = 0$ and $-(\Sigma_{11})^{-1} = 0$. Since the correlation between $X$ and $Y$ is less than 1, one can show that the regression coefficient of $X$ on $Y$ approaches to 0 when the variance of $X$ approaches to $\infty$, i.e., $(\Sigma_{11})^{-1}\Sigma_{12} = 0$. Thus, we can represent this case using a partially swept matrix as follows:

$$M(\vec{X},Y) = \begin{bmatrix} 0 & \mu_2 \\ 0 & 0 \\ 0 & \Sigma_{22} \end{bmatrix}. \quad (7)$$

**Linear Equations**: Suppose $X$ and $Y$ are row vectors of normal variables with $Y = XA + b$, where $A$ and $b$ are the appropriate constant matrices. Since $X$ is an independent vector in the equation, we are completely ignorant about it. Therefore, $\mu_1(\Sigma_{11})^{-1} = 0$ and $-(\Sigma_{11})^{-1} = 0$. On the other hand, given $X = 0$, $Y$ is completely determined to be $b$. Thus, given $X = 0$, the conditional mean of $Y$ is $b$ and the conditional variance is 0. In addition, the regression coefficient matrix is $A$. Thus, the linear equations can be represented as a partially swept moment matrix:

$$M(\vec{X},Y) = \begin{bmatrix} 0 & b \\ 0 & A \\ A^T & 0 \end{bmatrix}. \quad (8)$$

The knowledge to be represented in linear equations is very close to that in a proper LBF, except that the former assume a perfect correlation between $X$ and $Y$ while the latter do not. This observation is interesting; it characterizes the difference between partial ignorance and linear equations in one parameter — correlation.

**Linear Regression Models**: Suppose $X$ and $Y$ are two vectors of normal variables with $Y = XA + b + E$, where $A$ and $b$ are the appropriate coefficient matrices and E is an independent white noise satisfying $E \sim N(0, \Sigma)$. This linear regression model may be considered to consist of two pieces of knowledge, one is specified by the linear equation involving three variables $X$, $Y$, and E, and the other is a simple normal distribution of E, i.e., $E \sim N(0, \Sigma)$. Alternatively, one may consider it similar to a linear equation, except that, given $X = 0$, $Y$ is not completely determined to be $b$. Instead, its conditional mean is $b$ and the conditional variance is $\Sigma$. Therefore, the linear regression model may be represented as a partially swept matrix:

$$M(\vec{X},Y) = \begin{bmatrix} 0 & b \\ 0 & A \\ A^T & \Sigma \end{bmatrix}. \quad (9)$$

From representing the six special cases, we see a clear advantage of the moment matrix representation, i.e., it allows for a unified representation for seemingly diverse types of knowledge, including linear equations, joint and conditional distributions, and ignorance. The unification is significant not only for knowledge representation in artificial intelligence but also for statistical analysis and engineering computation. For example, the representation treats the typical logical and probabilistic components in statistics—observations, distributional assumptions, improper priors, and linear equation models—not as separate concepts, but as manifestations of a single concept. It allows one to see the inner connections between these manifestations and interplay them for computation.

### 2.2 MAKING INFERENCES

There are two basic operations for making inferences in expert systems: combination and marginalization. Intuitively, combination aggregates knowledge while marginalization projects knowledge to a focused domain. Making inferences involves combining the knowledge into a "joint" and marginalizing the joint to variables of interest.

**Marginalization.** Marginalization involves removing variables. When expressed as a moment matrix, marginalization is simply the restriction of a non-swept moment matrix to a sub-matrix corresponding to the remaining variables. For example, for the joint distribution $M(X, Y)$ in Equation (5), its marginal for $Y$, denoted by $M^{\downarrow Y}$, is:

$$M^{\downarrow Y}(X,Y) = \begin{bmatrix} \mu_2 \\ \Sigma_{22} \end{bmatrix}. \quad (10)$$

When removing a variable, it is important that the variable has not been swept on, i.e., it does not have an arrow hat above the variable. It is easy to see from Equation (6) that one can remove any or all variables in $Y$ from the partially swept matrix $M(\vec{X},Y)$ and still produce the correct result—either partially or fully swept matrix for the remaining variables. To remove a variable that has been already swept on, we have to reverse the sweeping. Suppose a moment matrix is partially swept as follows:

$$M(\vec{X},Y) = \begin{bmatrix} \overline{\mu}_1 & \overline{\mu}_2 \\ \overline{\Sigma}_{11} & \overline{\Sigma}_{12} \\ \overline{\Sigma}_{21} & \overline{\Sigma}_{22} \end{bmatrix}. \quad (11)$$

Then a reverse sweeping on $X$ is defined as follows:

$$M(X,Y) = \begin{bmatrix} -\overline{\mu}_1(\overline{\Sigma}_{11})^{-1} & \overline{\mu}_2 - \overline{\mu}_1(\overline{\Sigma}_{11})^{-1}\overline{\Sigma}_{12} \\ -(\overline{\Sigma}_{11})^{-1} & -(\overline{\Sigma}_{11})^{-1}\overline{\Sigma}_{12} \\ -\overline{\Sigma}_{21}(\overline{\Sigma}_{11})^{-1} & \overline{\Sigma}_{22} - \overline{\Sigma}_{21}(\overline{\Sigma}_{11})^{-1}\overline{\Sigma}_{12} \end{bmatrix}. \quad (12)$$

Note that forward and reverse sweepings are opposite operations. Liu (1999b) proves that a moment matrix will be recovered through a reverse sweeping followed by a forward sweeping on the same set of variables. It can be also recovered through a forward sweeping followed by a reverse sweeping.

**Combination.** According to Dempster's rule, the combination of continuous belief functions may be expressed as the intersection of focal elements and the multiplication of probability density functions. Liu (1996) applies the rule to LBFs in particular and obtains a formula of combination in terms of density functions. Later he proves a claim by Dempster (1990) and re-expresses the formula as the sum of two fully swept matrices (Liu 1999b). Assume two LBFs for the same vector of variables $X$ as follows:

$$M_1(X) = \begin{bmatrix} \mu_1 \\ \Sigma_1 \end{bmatrix}, \quad M_2(X) = \begin{bmatrix} \mu_2 \\ \Sigma_2 \end{bmatrix}. \quad (13)$$

Then their combination, in the potential term, is

$$M(\vec{X}) = \begin{bmatrix} \mu_1 \Sigma_1^{-1} + \mu_2 \Sigma_2^{-1} \\ -\Sigma_1^{-1} - \Sigma_2^{-1} \end{bmatrix}. \quad (14)$$

When applying Equation (14), if two matrices to be combined have different dimensions, one or both matrices must be vacuously extended (Kong 1986), i.e., assuming ignorance on the variables that are not present in each matrix and using Equation (7) to extend it.

In general, to combine two LBFs, their moment matrices must be fully swept. However, one may combine a fully swept matrix with a partially swept one directly if the variables of the former matrix have been all swept on in the later. We can use the linear regression model —$Y = XA + b + \mathrm{E}$ — to illustrate the property. As we mentioned, the regression model may be considered as the combination of two pieces of knowledge: one is specified by the linear equation involving three variables $X$, $Y$, and E, and the other is a simple normal distribution of E, i.e., E ~ $N(0, \Sigma)$. Let their moment matrices be the following:

$$M_1(\vec{X},\vec{\mathrm{E}},Y) = \begin{bmatrix} 0 & 0 & b \\ 0 & 0 & A \\ 0 & 0 & I \\ A^T & I & 0 \end{bmatrix}, \quad M_2(\vec{\mathrm{E}}) = \begin{bmatrix} 0 \\ -\Sigma^{-1} \end{bmatrix}. \quad (15)$$

Then the two matrices can be combined directly without sweeping $M_1$ on $Y$ first. The combination is as follows:

$$M(\vec{X},\vec{\mathrm{E}},Y) = \begin{bmatrix} 0 & 0 & b \\ 0 & 0 & A \\ 0 & -\Sigma^{-1} & I \\ A^T & I & 0 \end{bmatrix}. \quad (16)$$

If we apply a reverse sweeping on E and then remove E from the matrix in Equation (16), we will obtain the same representation of the regression model as Equation (9).

Note that the simple formula in Equation (14) unifies many seemingly diverse operations, including Bayesian conditioning, solving linear equations, multiplying probability distributions, and most importantly, integrating independent belief functions. For example, statistical inference on linear models and Kalman filtering can be reduced to the one of combining LBFs.

## 3 PORTFOLIO EVALUATION

In this section, we use LBFs to evaluate a simple portfolio of three gold mining stocks: $S_1, S_2$, and $S_3$. Each stock is affected by changes in the stock market ($M$), the gold price ($G$), and firm-specific factor $F_1$, $F_2$, and $F_3$, respectively. The firm specific factors include anything that affects the firm, not already included in the other factors. We are interested in the return distribution of the portfolio ($P$). This example is small enough to illustrate the computations, yet it includes the features of a large class of portfolios. It builds on the traditional framework of multi-factor models, where the return on a stock is represented as a regression model:

$$r = \alpha + \beta_1 f_1 + \beta_2 f_2 + ... + \beta_k f_k + \varepsilon \quad (17)$$

where $r$ is the return on stock $i$, $\varepsilon$ is a random component of the return due to firm specific effects, $f_k$ is the return on factor $k$, and $\beta_k$ is the responsiveness of the stock $i$ to factor $k$. As we have seen, the multifactor model is a LBF and can be encoded as a partially swept matrix.

### 3.1 A VALUATION NETWORK

Figure 1 shows a graphical structure, called *valuation network* (VN), for representing the problem. There are two types of nodes in a VN. The elliptical nodes represent variables, and the rectangular nodes represent LBFs. A VN is essentially a bi-partite graph in which edges connect only between variable nodes and belief function nodes (Shenoy 1994). The edges between a rectangular node and elliptical nodes denote the domain of the belief function. In our example, there are nine variables represented by nine elliptical nodes. There are six belief functions represented by six rectangular nodes. Among them, four LBFs capture structural relations among the variables and the other two capture beliefs respectively on gold price and market level.

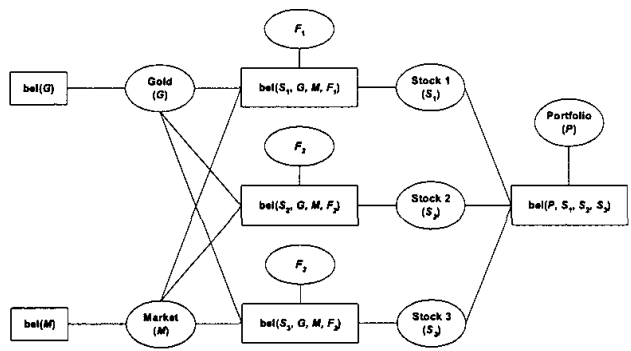

*Figure 1.* A Valuation Network for Portfolio Pricing

Multifactor models dictate the overall graphical structure. For example, the domain of $bel(S_1, G, M, F_1)$ consists of four variables: $S_1$, $F_1$, $G$, and $M$. It represents how the return of Stock 1 is related to the values of gold, market, and $F_1$. It may capture the linear regression model between $S_1$ and its ensuing factors $G$, $M$ and $F_1$. Two other structural functions in Figure 1 have a similar interpretation. The belief function $bel(P, S_1, S_2, S_3)$ bears on the variables $P$, $S_1$, $S_2$, and $S_3$. It shows how the portfolio return is related to the individual stock returns. Since a portfolio is a linear blend of individual assets (Tobin 1959), portfolio return is deterministically a weighted average of stock returns. Thus, $bel(P, S_1, S_2, S_3)$ could be a linear equation that links $P$ with its constituent stocks: $S_1, S_2$, and $S_3$. In sum, all the four structural belief functions in Figure 1 are special cases of a LBF and can be represented as partially swept moment matrices.

A LBF on an individual factor represents a belief on the true value of the factor. It takes one of many alternative forms. It could represent a distribution assumption as we typically make about residual variables. It could represent a subjective belief that, for example, the gold price is 34 dollars on average with standard deviation 2 dollars. It could represent an observation that the factor takes on a certain value. It could represent an empirical distribution obtained from historical data. It could also be a vacuous belief function, meaning we are ignorant about the factor. Regardless which form it eventually takes, it is a LBF represented as a moment matrix.

It is important to note that the number of belief functions depends on the number of available pieces of knowledge regarding the variables. Therefore, there can be more or less than six belief functions in the example. First, in the extreme case when there is no knowledge on any variable, there will be no belief functions. We can always represent such a situation using a vacuous belief function for all variables, but this is not necessary. Second, in some situations, we may have multiple pieces of knowledge regarding one set of variables. For example, we may have two independent sources of knowledge both about the price of gold ($G$). In this case, we will represent this situation using two rectangular nodes, instead of one as in Figure 1, connecting to the elliptical node $G$. Third, there may exist knowledge about a set of variables that are not connected by any belief function node in Figure 1. For example, there may be a piece of evidence justifying a relationship between gold price ($G$) and market return ($M$). Note that a multifactor model typically assumes the independence of $G$ and $M$ for validating related statistical techniques. Such assumptions do not have to be held in the belief function framework. Instead, what is critical to our approach is that each belief function entering into the network is based on a distinct piece of evidence and knowledge, i.e., there is no double counting of uncertain knowledge/evidence.

### 3.2 SPECIFYING LINEAR BELIEF FUNCTIONS

Now let us describe the details of the LBFs in Figure 1. To be consistent with traditional asset pricing models such as the arbitrage pricing theory (Ross 1976), we measure the variables as rates of return or change. Thus, $P$, $S_1, S_2$, and $S_3$ are respectively the rate of return of the portfolio and each of the individual gold stocks. $M$ and $G$ are the rates of return of the overall stock market and the gold sector. $F_1$, $F_2$, and $F_3$ are the noise terms associated with individual firm specific factors and are also measured as rates of change.

Let us first specify the LBFs $bel(G)$ and $bel(M)$. Each of the two functions has a single variable in its domain. Their role is similar to the prior distributions in a Bayes-ian probability model. However, unlike a Bayesian model, if we do not have any information regarding a variable, we can simply leave it out of the representation. We illustrate the construction of $bel(G)$ with an example. One kind of information that frequently affects gold prices is news that a central bank is selling a large amount of gold. Based on historical data or personal experience, assume that the volume of the transaction could negatively impact the gold price by 5% on the average. However, the actual rate of change could vary with standard deviation 2%. We represent this piece of information as moment matrix:

$$M(G) = \begin{bmatrix} -0.05 \\ 0.02^2 \end{bmatrix}. \quad (18)$$

In addition, let us assume there is a piece of breaking news that, for example, China is joining the WTO. In this case, nobody understands exactly the impact of the news on the market because there are no historical data. However, it sparks many speculations. One of them suggests that the news could boost the stock market by 10% on the average with a wide spread of 8%, meaning there is still a 11% probability that the news will have a negative impact on the stock market. We can represent the speculation as another moment matrix:

$$M(M) = \begin{bmatrix} 0.10 \\ 0.08^2 \end{bmatrix}. \quad (19)$$

Belief functions for firm specific factors can be similarly specified. For now, we assume that they are vacuous and simply ignore them. When we receive evidence about these firm specific factors, we can update these beliefs accordingly.

Table 1. Multiple Regression Result

| Stock | $\alpha_k$ | $\beta_{1k}$ | $\beta_{2k}$ | St. Error |
|---|---|---|---|---|
| 1 | 0.03 | 0.60 | 0.40 | 0.08 |
| 2 | 0.03 | 0.45 | 0.25 | 0.04 |
| 3 | 0.03 | 0.50 | 0.30 | 0.05 |

Next, let us specify structural belief functions that describe how each of the three gold stocks is related to the gold, market and the corresponding firm specific factor. Suppose we base on historical information and perform linear regressions according to the multifactor model in Equation (17). The regression model of Stock $k$ given the return on the market ($M$) and the return on gold ($G$) is given by $S_k = \alpha_k + \beta_{1k}G + \beta_{2k}M + F_k$, where $F_k$ is the firm specific residual term, $k = 1, 2, 3$. Assume Table 1 summarizes the result. According to the table, the regression model for Stock 1 is $S_1 = 0.03 + 0.6G + 0.4M$ with a standard error 0.08. It means that, given $G = 0$ and $M = 0$, the average return of Stock 1 is 0.03, which may be a riskless rate of return. The standard error term suggests that actual rate of return varies with a standard deviation 8%. Thus, the regression model can be represented as follows:

$$M(S_1, \vec{G}, \vec{M}, \vec{F_1}) = \begin{bmatrix} 0.03 & 0 & 0 & 0 \\ 0 & 0.60 & 0.40 & 1.0 \\ 0.60 & 0 & 0 & 0 \\ 0.40 & 0 & 0 & 0 \\ 1.0 & 0 & 0 & -0.08^{-2} \end{bmatrix}. \quad (20)$$

Note that here we employed Equation (16) to represent a linear regression. There are two cases in which we may want to make residual terms explicit: (1) when one currently has non-vacuous knowledge about them; and (2) when one wants to have the flexibility to enter additional information on the residuals later. The VN in Figure 1 makes all firm specific factors explicit and thus our matrix representation corresponds to it. The other two regression models can be similarly represented.

Finally, we describe the function linking the three stocks with the portfolio. This relationship is a deterministic one depending on the number of shares of each stock that constitutes the portfolio. Assume that 20% of the portfolio value is Stock 1, 70% is Stock 2, and 10% is Stock 3. Then, $P$ is simply the weighted average of the returns of the three stocks, which can be represented as follows:

$$M(P, \vec{S_1}, \vec{S_2}, \vec{S_3}) = \begin{bmatrix} 0 & 0 & 0 & 0 \\ 0 & 0.2 & 0.7 & 0.1 \\ 0.2 & 0 & 0 & 0 \\ 0.7 & 0 & 0 & 0 \\ 0.1 & 0 & 0 & 0 \end{bmatrix}. \quad (21)$$

### 3.3 MAKING INFERENCES

Once we have represented all the knowledge related to the stocks, we can make inferences about each one and the overall portfolio by first combining all LBFs using Equation (14), and then marginalizing the joint to a particular variable of interest using Equation (10). In our model, we are interested in computing the marginal of the joint belief function for the portfolio variable $P$ and for each individual stock variable.

When there are a large number of variables, it may be more efficient to compute the marginal of the joint without explicitly computing the joint using a local computation technique in the case of discrete probabilities (Lauritzen and Spiegelhalter 1988), of discrete belief functions (Kong 1986), and of normal distributions. Liu (1999b) shows that the combination and marginalization of LBFs satisfies the three axioms of Shenoy and Shafer (1990) and therefore justifies the feasibility of doing local computations over LBFs. By following the Shafer-Shenoy architecture (Shafer 1996), he also shows a detailed implementation including the construction of a join-tree, the propagation of messages across the join-tree, and computation of marginals by gathering the messages. In this subsection we follow the same architecture to compute the marginals of interest. In particular, we first marginalize the firm-specific factors from the LBFs $bel(S_i, G, M, F_i)$. Then we combine the results with LBFs in Equations (18-19). Finally, we marginalize the result to the three stock variables and combine the marginal with Equation (21). We then apply reverse sweepings to obtain the marginals for each individual stock and the portfolio as a whole as in Table 2.

Table 2. Moment Matrix $M(P, S_1, S_2, S_3)$

| 0.0343 | 0.0400 | 0.0325 | 0.0350 |
|---|---|---|---|
| 0.0017 | 0.0021 | 0.0017 | 0.0009 |
| 0.0021 | 0.0076 | 0.0007 | 0.0009 |
| 0.0017 | 0.0007 | 0.0021 | 0.0006 |
| 0.0009 | 0.0009 | 0.0006 | 0.0032 |

According to Table 2, the prediction is as follows: the rate of return for Stock 1 is 4% with standard deviation 8.7%, for Stock 2 is 3.3% with a standard deviation 4.6%, for Stock 3 is 3.5% with standard deviation 5.7%, and for the overall portfolio is 3.4% with standard deviation 4.1%. Note that, in our example, we assumed the weights of the portfolio. Actually, based on the mean and covariance matrix of the three stocks shown in Table 2, one may also compute an optimal portfolio, i.e., the optimal allocation of wealth on the three stocks. Assume a risk-averse mean-variance investor attempts to maximize Sharpe ratio (Gibbons *et al.* 1989). Then the optimal weights of the three stocks will be 0.13, 0.53, and 0.34 respectively.

Table 3. Updated Moment Matrix $M(P, S_1, S_2, S_3)$

| 0.0753 | 0.0908 | 0.0706 | 0.0774 |
|---|---|---|---|
| 0.0016 | 0.0020 | 0.0016 | 0.0008 |
| 0.0020 | 0.0074 | 0.0006 | 0.0008 |
| 0.0016 | 0.0006 | 0.0020 | 0.0005 |
| 0.0008 | 0.0008 | 0.0005 | 0.0031 |

### 3.4 ENTERING ADDITIONAL EVIDENCE

Besides being a device for integrating evidence from independent sources, the LBF approach may be also used to dynamically update predictions on asset performance when additional information becomes available. For example, consider a report that Asian jewelry makers are entering the gold market to stockpile inventory. Considering the size of these jewelry makers and the amount of gold they will stockpile, one estimates that the gold price will rise at a rate between 3~5%. For simplicity, let us assume 3~5% is the 95% confidence interval for the actual gold price. Then we can present the information as a moment matrix for $G$ with mean 0.04 and standard deviation 0.005. To integrate the evidence, we can simply sweep the matrix and combine it with other equations to compute the impact of the new evidence. The final result is an updated moment matrix in Table 3. As expected, the new evidence significantly boosts the predicted rates of return of all stocks and the portfolio. It also slightly reduced the standard deviation of their distributions, meaning the prediction becomes less uncertain. By using the new mean and covariance matrix, we compute the weights in an optimal portfolio of the three gold stocks to be 0.14, 0.52, and 0.34 respectively.

## 4 CONCLUSION

In this paper, we proposed a new knowledge-based approach to evaluating portfolio performance in a framework of linear belief functions. We proposed a unified representation device—swept moment matrices—to represent knowledge, including normal distributions, statistical observations, ignorance, linear equations, and linear regression models. We simplified the rules of marginalization and combination of linear belief functions. As we showed, the marginalization is simply the projection of a partially swept matrix by removing a variable, if it has not been swept on, and its corresponding elements from the matrix. The combination of two linear belief functions is simply the addition of their fully swept moment matrices, or a partially swept matrix with a fully swept one, if the variables of the later have all been swept on in the former.

We showed how to encode multifactor models, linear combination of stocks, speculations, historical observations, and ignorance into linear belief functions. We built a valuation-based system using a portfolio of three stocks, each of which is affected by three factors—the overall stock market, the gold market, and firm specific effects. We showed how to make inferences using the system and how to update the belief by entering additional information into the system. We calculated the weights of an optimal portfolio by maximizing the Sharpe ratio.

Our approach to portfolio evaluation has many advantages over a Bayesian one. First, it represents all knowledge directly and naturally in a unified representation — moment matrices. In contrast, a Bayesian approach has to translate each piece of market evidence into a prior on a pricing model (Baks *et al.* 2001, Pástor 2000, Pástor and Stambaugh 1999). It represents different information using different devices such as priors, posteriors, observations, and equations. Second, our approach allows multiple views, coherent or contradicting, expressed on a same variable. It combines multiple asset pricing models along with other knowledge. Third, our approach represents ignorance without using non-informative priors, which Fisher (1959) regarded as "completely bogus" and Shafer (1976) considered confusing lack of belief with disbelief.

Our approach is superior to the ones proposed in Shenoy and Shenoy (1999, 2002). Among many reasons, the most important is that it does not discretize continuous variables and represent linear equation or regression models as subsets of ordered pairs of values. Note that the discretization procedure not only introduces approximation errors to knowledge representation but also makes computation complex and the modeling of a large-size portfolio intractable. For example, assume a portfolio consists of 500 individual stocks. Even using local computation, the largest sample space in computation has at least $2^{500}$ val-ues, which is too large even for powerful workstations. In contrast, our new approach does not have this problem. By using moment matrices, a computation involving 500 variables has to deal with only 125,750 distinct values, including 500 means and 125,250 variance and covariance values. The complexity becomes manageable.